\documentclass{article}
\usepackage{algorithm} 
\usepackage{algpseudocode} 
\usepackage{PRIMEarxiv}
\usepackage[table,xcdraw]{xcolor}
\usepackage[utf8]{inputenc} 
\usepackage[T1]{fontenc}    
\usepackage{hyperref}       
\usepackage{url}            
\usepackage{booktabs}       
\usepackage{amsfonts}       
\usepackage{nicefrac}       
\usepackage{microtype}      
\usepackage{lipsum}
\usepackage{fancyhdr}       
\usepackage{graphicx}       
\graphicspath{{media/}}     
\usepackage{adjustbox}
\usepackage{algorithm} 
\usepackage{algpseudocode}
\usepackage[section]{placeins}
\usepackage{amsmath}
\usepackage{amssymb}
\usepackage{booktabs}
\usepackage{multirow}
\usepackage{multirow,diagbox,tabularx,blindtext}
\usepackage{threeparttable}
\usepackage{color}
\usepackage{float}
\usepackage{tabularray}
\usepackage{longtable}
\usepackage[authoryear]{natbib}

\bibpunct{(}{)}{,}{a}{}{;}
\usepackage{graphicx}
\usepackage{subcaption}

\usepackage{lineno}
\title{Leveraging Large Language Models with Chain-of-Thought and Prompt Engineering for Traffic Crash Severity Analysis and Inference

}

\author{
  Hao Zhen, Yucheng Shi, Yongcan Huang, Jidong J. Yang*, Ninghao Liu\\
  College of Engineering\\
 University of Georgia, Athens, GA, USA\\
  \texttt{\{Hao.Zhen, Yucheng.Shi, Yongcan.Huang, Jidong.Yang, Ninghao.Liu\}@uga.edu} \\
}

\begin{document}
\maketitle

\begin{abstract}

Harnessing the power of Large Language Models (LLMs), this study explores the use of three state-of-the-art LLMs, specifically GPT-3.5-turbo, LLaMA3-8B, and LLaMA3-70B, for crash severity inference, framing it as a classification task. We generate textual narratives from original traffic crash tabular data using a pre-built template infused with domain knowledge. Additionally, we incorporated Chain-of-Thought (CoT) reasoning to guide the LLMs in analyzing the crash causes and then inferring the severity. This study also examine the impact of  prompt engineering specifically designed for crash severity inference. The LLMs were tasked with crash severity inference to: (1) evaluate the models' capabilities in crash severity analysis, (2) assess the effectiveness of CoT and domain-informed prompt engineering, and (3) examine the reasoning abilities with the CoT framework. Our results showed that LLaMA3-70B consistently outperformed the other models, particularly in zero-shot settings. The CoT and Prompt Engineering techniques significantly enhanced performance, improving logical reasoning and addressing alignment issues. Notably, the CoT offers valuable insights into LLMs' reasoning processes, unleashing their capacity to consider diverse factors such as environmental conditions, driver behavior, and vehicle characteristics in severity analysis and inference.

\end{abstract}


\section{Introduction}

Traffic safety research plays a crucial role in enhancing road safety by examining the root causes of accidents, identifying hazardous behaviors or factors, and proposing effective countermeasures \citep{mannering2020big}. Despite advancements in vehicle safety, enhancements in road design, and the implementation of various policies, traffic safety remains a significant challenge. One important aspect of road safety research is understanding contributing factors leading to different crash severity outcomes, which is essential for mitigating crash consequences. 

The challenge of traffic accident modeling stems from their multifaceted nature, involving intricate interplay among diverse factors, such as human behavior, vehicle dynamics, traffic conditions, environmental factors, and roadway characteristics. Traffic safety research has been primarily focused on understanding causality using observational data, due to the impracticality of conducting controlled experiments in this field \citep{mannering2020big}. Traditional statistical and econometric methods have long been studied in traffic safety domain \citep{golob2003relationships, eluru2007joint, lord2010statistical, savolainen2011statistical, karlaftis2002effects} for causality understanding. These classic methods suffer from several limitations, including constraints imposed by specific functional forms and distributional assumptions \citep{mannering2020big}, as well as subtle confounding effects, also referred as heterogeneous treatment effects \citep{mannering2020big, zhang2022inferring, pervez2022exploring}, which often lead to an incomplete or misleading understanding of influencing factors. 

Another limitation of statistical and econometric methods lies in the fact that they were designed around and can only consume structured data with traditional numeric or categorical coding and  a limited number of features. These methods can not effectively handle unstructured textual data or passages of  narratives. Due to recent advancements in AI and the abundance of narrative data captured in crash reports, natural language processing (NLP) methods have been applied in text mining of crash narratives \citep{goh2017construction, zhang2020identifying, das2021applying}. 
In previous works, researchers are required to collect a large amount of high-quality, labeled crash reports for model training. However, this process is time-consuming and costly. Additionally, low-quality training data and poorly chosen training parameters can lead to undesirable performance. In contrast, large language models (LLMs) offer a distinct advantage by leveraging their immense knowledge acquired from extensive pre-training with vast datasets, effectively addressing these challenges. 
Motivated by their superior capability to comprehend and generate human-like text, this study aims to investigate whether LLMs can process complex and unstructured data in traffic safety domain to enable elaborate case-specific analysis.

Despite the release of numerous LLMs, represented by the GPT family~\citep{openai2023gpt35} and LLaMA family ~\citep{llama3modelcard}, their ability for traffic crash analysis and reasoning remains unexplored. 
Applying LLMs to crash analysis presents two major challenges: (1) it requires LLMs to fully understand the domain knowledge and potential causality behind crash events. However, LLMs, typically built on transformer architectures, are often regarded as “black-box” models, making it difficult to interpret their decision-making processes.
(2) while LLMs possess extensive real-world knowledge acquired from the pre-training stage, they are not specialized in analyzing textual data in crash reports. This creates an alignment gap between the model’s original intent and the specific requirements of this specific task.  

To address the first challenge, we propose to leverage the Chain-of-Thought (CoT) technique to enhance the reasoning capabilities of LLMs~\citep{wei2022chain}. This technique guides the model through a structured reasoning process, helping it better understand the detailed knowledge and causality behind crash data. Additionally, the CoT approach provides explainable reasoning steps for each intermediate result, making the model’s decisions more transparent and easier to interpret. By incorporating CoT, we aim to improve the LLMs’ performance in crash severity modeling, leveraging their capability to effectively process the complex and diverse data relevant to crash analysis.

To address the second challenge, we propose to use prompt engineering (PE) and few-shot learning (FS) to better align the LLMs with the specific requirements of the target task: crash severity modeling and analysis. PE can tailor the input prompts to guide the LLMs toward more relevant and reliable analysis, while few-shot learning can provide the models with specific examples to improve their understanding and performance in the subject domain. By combining these techniques, we aim to bridge the alignment gap and enhance the models' ability to effectively analyze textual descriptions in crash reports.

To demonstrate the efficacy of our approach, we explore three state-of-the-art LLMs, specifically GPT-3.5-turbo, LLaMA3-8B, and LLaMA3-70B, for crash severity inference, framing it as a multi-class classification task. In our experiments, we utilized textual narratives derived from crash tabular data as input for crash severity analysis with LLMs. Additionally, we incorporated CoT to guide the LLMs in analyzing potential crash causes and subsequently inferring severity outcome. We also examine prompt engineering specifically designed for crash severity inference. 
We task LLMs with crash severity inference to (1) assess the ability of LLMs for crash severity analysis; (2) evaluate the effectiveness of CoT and domain-informed PE; and (3) examine the reasoning ability within the CoT framework.

The experimental setup involves several strategies, including plain zero-shot and few-shot settings, zero-shot with Chain-of-Thought (ZS\_CoT), zero-shot with prompt engineering (ZS\_PE), zero-shot with both prompt engineering and Chain-of-Thought (ZS\_PE\_CoT), and few-shot with prompt engineering (FS\_PE). The LLMs evaluated include GPT-3.5-turbo, LLaMA3-8B, and LLaMA3-70B, with specific hyperparameters to ensure consistent and reliable results. We compare the performance of these models and settings to determine the most effective approach for the crash severity inference task.

\section{Methods}

\subsection{Model Descriptions} 
In our approach, we leverage the reasoning ability of LLMs for domain-specific (i.e., traffic safety) analysis and modeling. Specifically, we utilize two state-of-the-art LLMs as our base models, including GPT-3.5-turbo~\citep{openai2023gpt35} and LLaMA-3~\citep{llama3modelcard}.
Both models are auto-regressive language models, which generate text by predicting the next word or subword in a sequence based on the previous words or subwords, one step at a time. This approach allows the model to create coherent and contextually relevant text. In this auto-regressive setting, the joint probability is expressed as the product of sequential conditional probabilities in Eq. 1:

\begin{equation}
    P(x_1, x_2, \ldots, x_n) = \prod_{i=1}^n P(x_i \mid x_1, x_2, \ldots, x_{i-1}), \label{eq:prob}
\end{equation}

where $P(x_i \mid x_1, x_2, \ldots, x_{i-1})$ represents the conditional probability of the $i$-th word given the preceding words. This auto-regressive modeling framework, designed to handle large context windows and trained on extensive datasets, empowers the model to generate fluent and context-aware sequences. Some details about these two models are provided below for direct reference.

GPT-3.5-turbo~\citep{openai2023gpt35}, developed by OpenAI, is designed for a variety of applications, including advanced text generation, coding assistance, and more. Trained on a vast corpus of internet text, including diverse sources such as books and websites, it has billions of parameters, allowing for nuanced understanding and generation of text. In our experiments, we use the \textit{gpt-3.5-turbo-0125} version. 

LLaMA-3~\citep{llama3modelcard}, developed by Meta, is another state-of-the-art LLM designed for efficient and scalable language understanding. It is pre-trained on over 15T tokens that cover diverse internet text sources. LLaMA-3 is available in sizes of 8 billion and 70 billion parameters, making it flexible for various applications.

Both models employ techniques like supervised fine-tuning (SFT) and Reinforcement Learning with Human Feedback (RLHF) to align their outputs with human preferences, ensuring the model is helpful and safe~\citep{wu2023language}. Specifically, SFT fine-tunes a pre-trained model on specific datasets with human instructions, ensuring it understands relevant vocabulary, patterns, and knowledge, thereby improving accuracy and relevance. RLHF, on the other hand, refines the model using feedback from human experts, allowing it to adapt to complex real-world scenarios and prioritize critical safety aspects, enhancing both adaptability and reliability. 

\subsection{In-context learning}
In-context learning is a promising approach for traffic safety analysis and modeling, where a LLM learns to perform a specified task by observing examples of the task within the input context, without any explicit fine-tuning or gradient updates~\citep{radford2019language}. The LLM leverages its pre-existing knowledge and the provided examples to generate appropriate inference for new, unseen instances of the task. In-context learning encompasses both zero-shot and few-shot learning~\citep{brown2020language}, with the number of examples ranging from zero to a few. In the context of traffic safety analysis and modeling, in-context learning can be applied to various tasks, such as crash severity inference.

Zero-shot learning in traffic safety analysis refers to the setting where the model is expected to perform a task without any examples or explicit training for that specific task. The model relies solely on its pre-existing knowledge to make inference. While few-shot learning involves providing the model with a few examples of the target task before asking it to perform the same task on new instances. 
The model learns from these few examples and adapts its behavior accordingly. For instance, in crash severity inference, the model may be provided with a few examples of crashes and their corresponding severity outcomes. 
The model then learns from these examples and reasons the severity outcome of a new, unseen crash. Few-shot learning allows the models to quickly adapt to new tasks or scenarios with minimal training data for diverse real-world applications~\citep{yin2020meta}.

\subsection{Prompt engineering (PE)}
Prompt engineering is a technique for crafting and refining input prompts for LLMs to enhance their performance on specific tasks. It is akin to "asking the right question (prompt) to get the desired answer (response)." 
For traffic safety analysis, carefully designed prompts can significantly enhance LLMs' ability to analyze complex scenarios and provide meaningful insights.
Specifically, in the context of crash severity inference, one critical category is 'Fatal accidents'. However, LLMs often exhibit a tendency to avoid assigning this label to relevant cases. This behavior stems from their alignment training, which generally encourages them to avoid discussing unpleasant subject or potentially unsafe topics ~\citep{wang2023aligning, shen2023large}. Such design constraints present a challenge in accurately inferring severe outcomes for traffic incidents.
To address this issue, we found it necessary to rephrase the original 'Fatal accidents' label using alternative terms. This "soft" approach allows us to maintain inference accuracy while respecting LLMs' aligned parameters. The specific modifications and their impacts on inference performance will be discussed in detail in the Experiments section.

\subsection{Chain-of-Thoughts (CoT)}

A notable distinction between LLMs and conventional machine learning algorithms is the capacity of foundation LLMs to process variably structured input data, specifically prompts, during the inference phase \citep{wei2022chain}. For inference, LLMs generally precede or prioritize the information provided in the input prompts over the large implicit knowledge gained from the pre-training stage. Consequently, this leads to clear, comprehensive content. This content could encompass domain-specific knowledge, contextual background, or detailed step-by-step reasoning guidance. By doing so, LLMs can be encouraged to disclose their reasoning processes during inference, thereby enhancing the explicability of their outputs. In this paper, we utilize the CoT technique to enable LLMs to generate step-by-step reasoning before providing the final answer, improving their performance on complex reasoning tasks.

Although it is widely recognized that LLMs excel at generating responses reminiscent of human conversation, they often have opacity issues in their reasoning processes. This opacity hinders users' ability to evaluate the trustworthiness of the responses, particularly in scenarios that necessitate elaborate reasoning.  

Formally, let $f_{\theta}$ be a language model and $X = \{(x_1, y_1), (x_2, y_2),..., (x_n)\}$ be an input prompt, where $(x_i, y_i)$ denotes the $i$-th example question-response pair. In a standard question-answering scenario, the model output is given by: $y_n = \arg\max_{Y} p_{\theta}(Y | x_1, y_1, x_2, y_2,..., x_n)$, where $p_{\theta}$ is the predicted probability of the language model $f_{\theta}$.
This setting, however, does not provide insights into the reasoning process behind the answer $y_n$. The CoT method extends this by including human-crafted explanations $e_i$ for each example, resulting in a modified input format $X = {(x_1, e_1, y_1), (x_2, e_2, y_2),..., (x_n)}$. The model then outputs both the explanation $e_n$ and the answer $y_n$:

\begin{equation}
(e_n, y_n) = \arg\max_{Y} p_{\theta}(Y |x_1, e_1, y_1, x_2, e_2, y_2,..., x_n).
\end{equation}
However, in practice, it is difficult to obtain sufficient high-quality explanation examples for crash severity classification, while low-quality explanations can harm the CoT performance. Therefore, following strategies in~\citep{kojima2022large}, we simply ask LLMs to think step by step to conduct traffic safety analysis and inference. Some template examples are presented in the Experiments section. 
Besides allowing for a more transparent and understandable interaction with LLMs, the CoT approach is also practically useful in several key aspects: (1) Reducing Errors in Crash Risk Assessment: By breaking down complex traffic scenarios, CoT can better understand and reason over specific cases, thus reducing errors in crash risk analysis~\citep{wei2022chain, qin2023chatgpt, zhang2023multimodal}. (2) Providing Adjustable Intermediate Steps for Crash Analysis: CoT enables the outlining of traceable intermediate steps within the crash analysis process. 
 
This can be helpful in identifying potential biases or errors in the model's reasoning and allow for more reliable crash risk assessment~\citep{lyu2023faithful}.
By leveraging the CoT approach in traffic safety analysis, we can enhance the interpretability and reliability  of crash risk assessment.

\section{Data}

In this section, we first discuss the dataset employed for the study. We then explain how we convert the crash tabular data to coherent descriptive narratives. Finally, we discuss our experimental settings and evaluation methods.

\subsection{Dataset}
Our empirical analysis utilizes data sourced from CrashStats data from Victoria, Australia spanning from 2006 to 2020. The crash database contains records of vehicles involved in crashes. A four-point ordinal scale is used to code the severity level of each accident, including: (1) non-injury accident, (2) other injury (minor injury) accident, (3) serious injury accident, and (4) fatal accident. Each sample denotes a vehicle involved in a crash with driver's information. After data prepossessing, the final dataset has an extremely low representation of non-injury accidents (only four instances), accounting for less than 0.001\%. Consequently, these four non-injury accidents are merged to the category of "Minor or non-injury accidents". As a result, the dataset contains 197,425 minor or non-injury accidents, 89,925 serious injury accidents, and 4,760 fatal accidents.  

The traffic accident attributes considered in our empirical study include crash characteristics, driver traits, vehicle details, roadway attributes, environmental conditions, and situational factors (see Table \ref{tab:variables}).

\begin{longtblr}[
  caption = {Traffic accident attributes},
  label = {tab:variables},
]{
  hline{1,52} = {-}{0.08em},
  hline{2} = {-}{},
}
\textbf{Variables}               & \textbf{Description}                                                                              \\
\textbf{Crash characteristics}   &                                                                                                   \\
ACCIDENT\_TYPE            & The type of accident.                                                                             \\
EVENT\_TYPE               & Type of incident event.                                                                           \\
VEHICLE\_1\_COLL\_PT      & Collision point on the first vehicle involved in the event.                                       \\
VEHICLE\_2\_COLL\_PT      & Collision point on the second vehicle involved in the event.                                      \\
OBJECT\_TYPE              & Object involved in the specific accident event.                                                   \\
DCA                       & The definitions for classifying accidents.                                                        \\
ACCIDENT\_MONTH           & The month in which the accident occurred, derived from "ACCIDENT\_DATE".                           \\
TIME\_PERIOD             & The period the accident occurred, derived from "ACCIDENT\_TIME".               \\
DAY\_OF\_WEEK             & The day of the week the accident occurred.               \\
LGA\_NAME                 & The name of local government areas.                                                                                         \\
REGION\_NAME              & The region where the accident occurred.                                                           \\
DEG\_URBAN\_NAME          & The type of urbanized area for the crash site.                                                  \\
\textbf{Driver characteristics}  &                                                                                                   \\
DRIVER\_SEX               & The sex of the driver.                                                                            \\
AGE\_GROUP                & The age group of the driver, derived from "DRIVER\_AGE".                                          \\
ROAD\_USER\_TYPE          & The role of the person was at the time of the accident.                                           \\
\textbf{Vehicle characteristics} &                                                                                                   \\
VEHICLE\_TYPE             & The type or category of vehicle.                                                                  \\
VEHICLE\_WEIGHT           & The weight or mass of the vehicle. The unit of measurement is kilograms.                          \\
NO\_OF\_WHEELS            & The number of wheels that the vehicle has.                                                        \\
SEATING\_CAPACITY         & The number of seats in the vehicle.                                                               \\
FUEL\_TYPE                & The type of fuel used by the vehicle.                                                             \\
VEHICLE\_AGE                     & The age of the vehicle when the accident occurred. \\
VEHICLE\_BODY\_STYLE      & The body type of the vehicle.                                                                     \\
TRAILER\_TYPE             & The type of trailer towed by the vehicle involved in the accident.                                \\
\textbf{Roadway attributes}      &                                                                                                   \\
ROAD\_TYPE                & Type of the highest priority road at the intersection or the road the accident occurred.           \\
ROAD\_GEOMETRY            & The layout of the road where the accident occurred.                                               \\
SPEED\_ZONE               & The speed zone at the location of the accident.                                                   \\
ROAD\_SURFACE\_TYPE       & The type of road surface: 1: Paved 2: Unpaved 3: Gravel 9: Not known.                             \\
ROAD\_TYPE\_INT           & The type or suffix of the intersecting road.                                                      \\
COMPLEX\_INT\_NO          & Whether or not the segment is part of a complex intersection.                                     \\
\textbf{Environmental factors}   &                                                                                                   \\
LIGHT\_CONDITION          & The light condition or level of brightness at the time of the accident.                           \\
SURFACE\_COND             & Road surface condition: dry, wet, muddy, snowy, icy, unknown.                                      \\
SURFACE\_COND\_SEQ        & Starts with 1 and incremented by 1 if more than one road surface condition.                       \\
ATMOSPH\_COND             & Atmospheric condition.                                                                             \\
ATMOSPH\_COND\_SEQ        & 1 and incremented by 1 if more than one atmospheric condition is entered.   \\
\textbf{Situational factors}       &                                                                                                   \\
HELMET\_BELT\_WORN        & Whether or not the person was wearing a helmet or seatbelt at   the time of the accident          \\
NO\_OF\_VEHICLES          & The number of vehicles involved in the accident.                                                \\
LAMPS                     & Whether the lamps or headlights for the vehicle were alight (on).                                 \\
VEHICLE\_MOVEMENT         & The actual movement of the vehicle before the accident.                                         \\
TRAFFIC\_CONTROL          & The type of traffic control measure in the location where the accident occurred.                  \\
NO\_PERSONS               & The number of people involved in the accident.                                                  \\
NO\_OCCUPANTS             & The number of occupants or people in the vehicle at the time of the accident.                     \\
SUB\_DCA                  & SUB\_DCA   code and description of the accident.                                                                  \\
SUB\_DCA\_SEQ             & Starts with 1 and is incremented by 1 if more than one sub\_dca is entered.    \\
DRIVER\_INTENT            & The intent of the driver initially.                                                                   
\end{longtblr}

\subsection{Textual narrative generation}

\begin{figure}[H]
    \centering
    \includegraphics[width=0.7\linewidth]{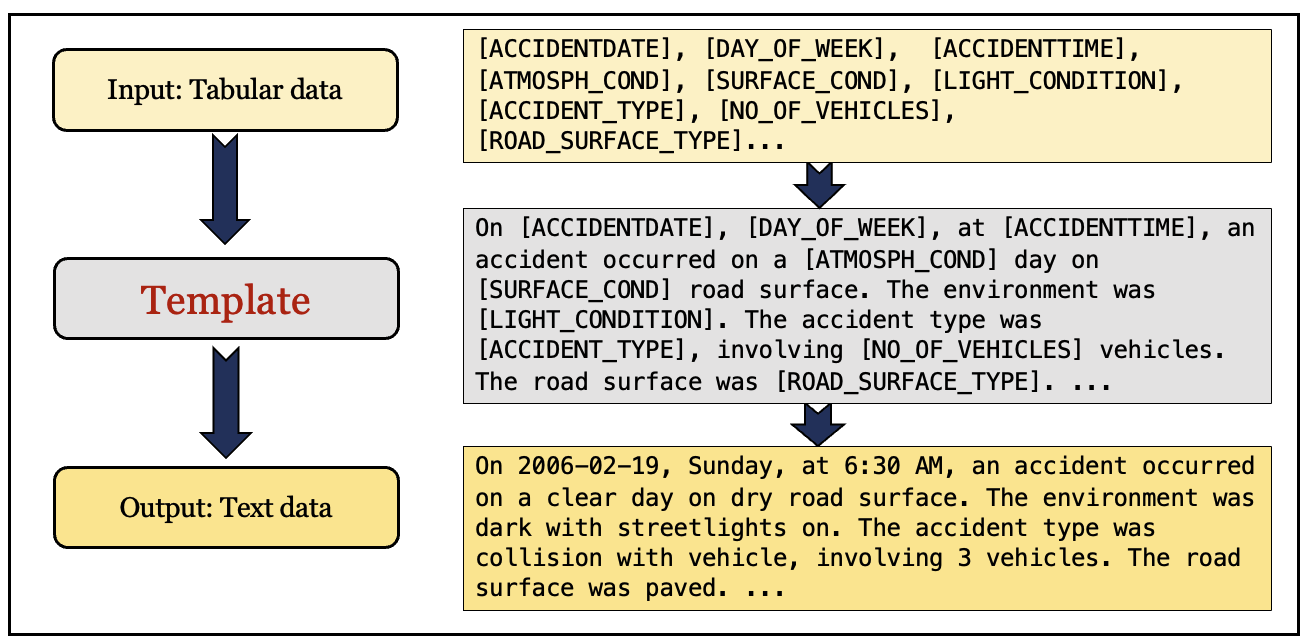}
    \caption{Textual narrative generation}
    \label{fig:tab2text}
\end{figure}

To get coherent, informative passages enriched with domain-specific knowledge, we use a simple yet effective template to convert the raw structured tabular data into detailed, human-readable textual narratives, encapsulating vital information about traffic accidents, which can be better consumed by LLMs. This process is depicted in Figure \ref{fig:tab2text}. 

The primary objective is to augment the applicability and relevance of tabular data as input for LLMs, facilitating more context-aware inference for tasks like accident severity assessment. Furthermore, road safety engineers can supplement it with established facts or domain-specific knowledge for further enhancement. For instance, "Children and elders are typically more vulnerable in accidents without seat belts." However, it is important to note that, for this study, we do not delve into the intricate design of scientific knowledge in traffic safety.

\section{Experiments}

\subsection{Experiments design}
In this work, we tackle the crash severity inference problem as a classification task. The inputs encompass various crash related attributes, including environmental conditions, driver characteristics, crash details, and vehicle features. The original data is in tabular format including categorical and numerical fields. We then transform the tabular data into consistent textual narrative with a simple template, detailed in the preceding subsection. The objective is to estimate severity outcomes of crashes with the state-of-the-art LLM models, such as GPT-3.5-turbo, LLaMA3-8B, and LLaMA3-70B.

LlaMA3s models are open-source foundation models, while the GPT model is a close-source and non-free model. Considering the cost and our goal of evaluating crash severity inference performance of foundation models, we randomly draw 50 samples from each of the three severity outcome categories, resulting in a total of 150 samples. These samples are used to demonstrate the potential of LLMs in enhancing crash analysis and reasoning.

\begin{table}[h]
\centering
\caption{Experiments}
\label{experiments}
\begin{adjustbox}{max width=\textwidth}
\begin{tabular}{lcccc}
\toprule
\multicolumn{5}{l}{\textbf{Experiments Setting}} \\
\midrule
 & Plain & Chain-of-Thought & Prompt Engineering & Prompt Engineering with Chain-of-Thought \\
Zero-shot & ZS & ZS\_CoT & ZS\_PE & ZS\_PE\_CoT \\
Few-shot & FS & / & FS\_PE & / \\
\midrule
\multicolumn{5}{l}{\textbf{Models}} \\
\midrule
& GPT-3.5-turbo & LLaMA3-8B & LLaMA3-70B & \\

Sampling Strategy & Greedy & Greedy & Greedy & \\
(temperature, top\_p) & (0, 0.01) & - & - & \\
\bottomrule
\end{tabular}
\end{adjustbox}
\end{table}

The strategies outlined in Table \ref{experiments} includes: Zero-shot and Few-shot settings coupled with different techniques. We use ZS and FS to denote the plain zero-shot and few-shot setting without prompt engineering and Chain-of-Thought. Other settings are zero-shot with Chain-of-Thought (ZS\_CoT), zero-shot with prompt engineering (ZS\_PE), zero-shot with prompt engineering and Chain-of-Thought (ZS\_PE\_CoT), and few-shot with prompt engineering (FS\_PE). It worth noting that some literature treats chain-of-thought (CoT) as a special form of Prompt Engineering (PE). In this paper, we make distinction between PE and CoT to highlight the advantages of CoT over a basic PE approach in the context of traffic safety analysis.

With these experiments, we aim to determine: 1) the accident severity inference performance of LLMs in a plain zero-shot setting; 2) whether CoT enhances the performance through its reasoning process (ZS\_CoT vs ZS; ZS\_PE vs ZS\_PE\_CoT); 3) whether PE improves performance in zero-shot and few-shot settings (ZS\_PE vs ZS; FS\_PE vs FS); 4) whether few-shot learning boost performance compared to the zero-shot setting (ZS vs. FS).  Accordingly, we tested six prompts to automatically infer the severity outcome of crashes. The details of these prompts are presented in the following section.

The experiments were conducted using GPT-3.5-turbo, LLaMA3-8B, and LLaMA3-70B. For GPT-3.5-turbo and LLaMA3, the hyperparameters are configured with temperature = 0 and top\_p = 0.0001 for crash severity inference, aiming to produce consistent and reliable results through greedy decoding. Additionally, the LLaMA3 models are set with $do\_sample = FALSE$ to generate deterministic outputs, ensuring reproducibility.

\subsection{Prompts for LLMs}
In this section, we explained in detail how we design the prompts for different experiments in Table \ref{experiments}.

\subsubsection{Zero-shot}

\begin{figure}[H]
    \centering
    \includegraphics[width=0.8\linewidth]{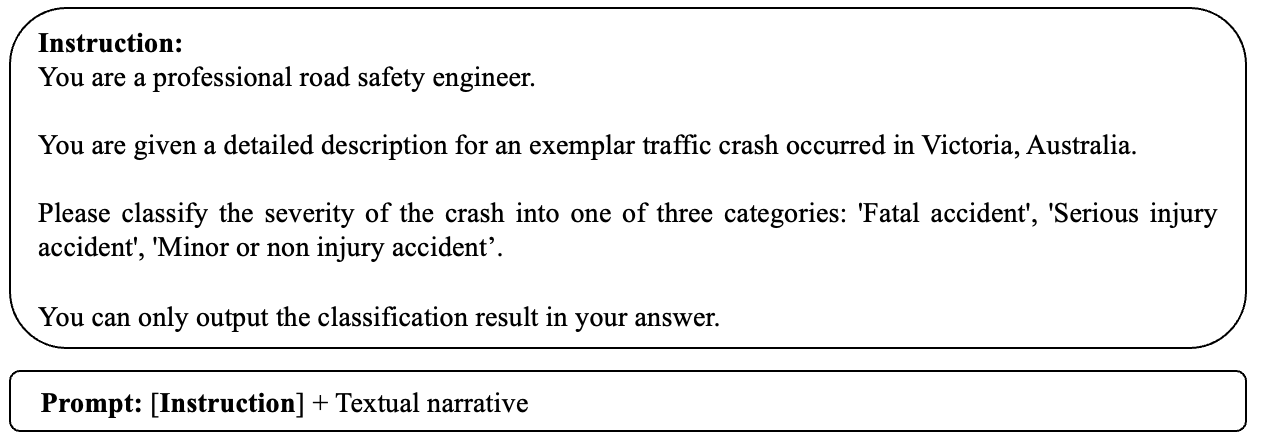}
    \caption{Zero-shot (ZS)}
    \label{fig:ZS}
\end{figure}

The prompt designed for plain zero-shot setting is demonstrated in Figure \ref{fig:ZS}. The provided prompt tasks each LLM as a professional road safety engineer with classifying the severity of a traffic crash in Victoria, Australia, based on a detailed description of a crash. The engineer is required to categorize the crash into one of three specified categories: 'Fatal accident', 'Serious injury accident', or 'Minor or non-injury accident'. The engineer's response is restricted to outputting only the classification result, ensuring a focused and objective assessment. This prompt is designed to elicit a precise evaluation of the crash severity outcome, leveraging the knowledge of engineer's expertise in road safety and crash analysis.

\begin{figure}[H]
    \centering
    \includegraphics[width=0.8\linewidth]{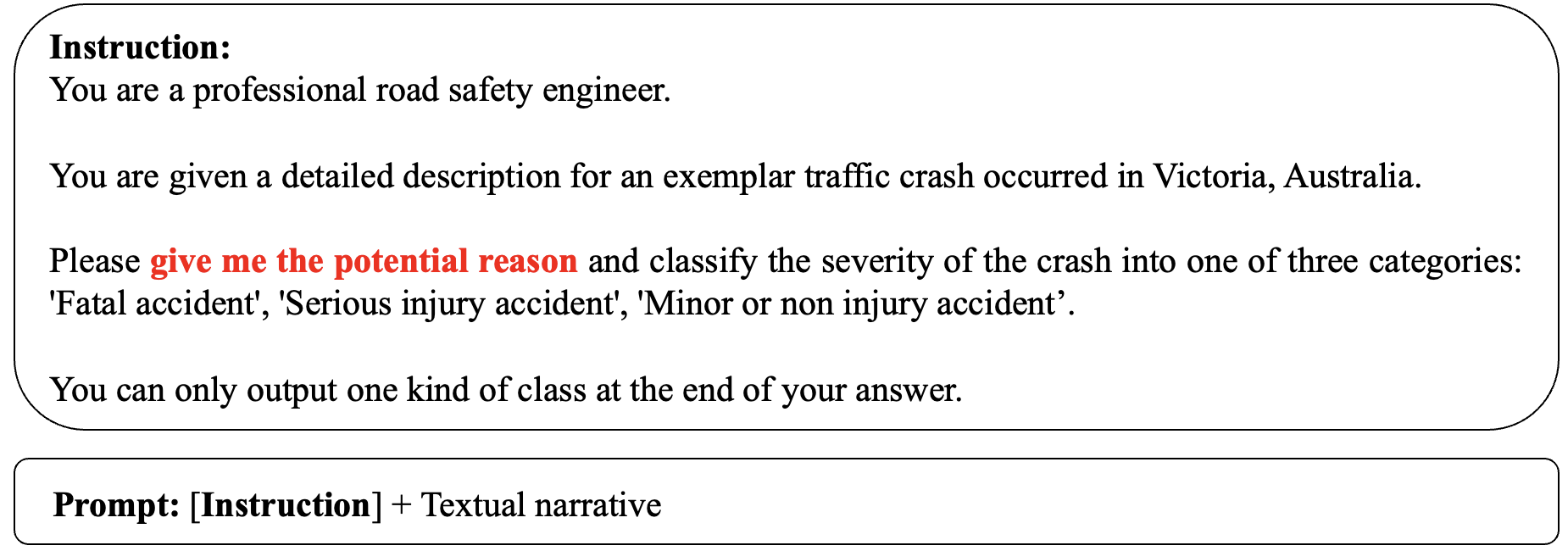}
    \caption{Zero-shot with CoT (ZS\_CoT)}
    \label{fig:zero-shot with CoT}
\end{figure}

The prompt designed for zero-shot with CoT setting is demonstrated in Figure \ref{fig:zero-shot with CoT}. This prompt leverages a CoT approach, encouraging the LLM to serve as an engineer to methodically reason through the details of the accident to determine both the cause and the severity outcome, thereby ensuring a comprehensive and structured assessment based on LLMs' knowledge. The difference between ZS\_CoT and plain ZS is highlighted in red color in Figure \ref{fig:zero-shot with CoT}.

\begin{figure}[H]
    \centering
    \includegraphics[width=0.8\linewidth]{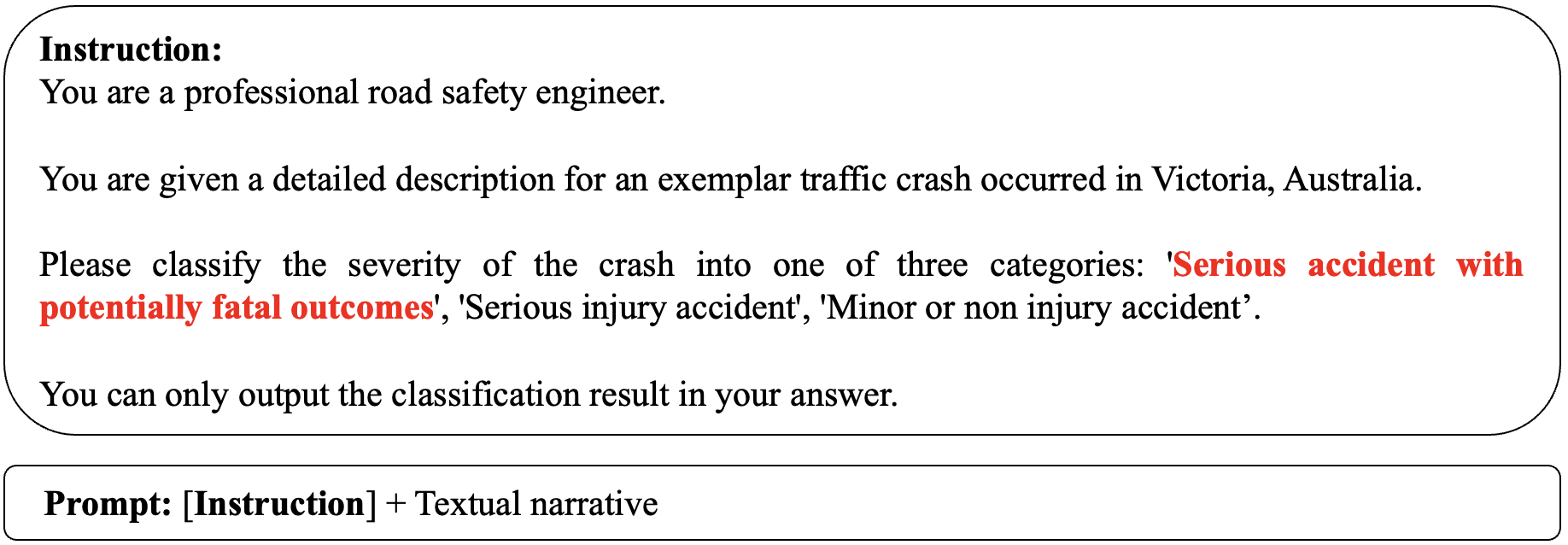}
    \caption{Zero-shot with prompt engineering (ZS\_PE)}
    \label{fig:ZSPE}
\end{figure}

The prompt designed for zero-shot with prompt engineering setting is demonstrated in Figure \ref{fig:ZSPE}. The provided prompt instructs LLM to serve as a professional road safety engineer to classify the severity outcome of a traffic crash, using a descriptively modified set of categories (see the revised class description in red in Figure \ref{fig:ZSPE}.) to accommodate alignment constraints in LLMs. The engineer must categorize the crash into one of three revised descriptive labels: 'Serious accident with potentially fatal outcomes', 'Serious injury accident', or 'Minor or non-injury accident'. The prompt explicitly requires the engineer to output only the classification result.

This rephrasing aims to preserve classification accuracy while adhering to the alignment parameters of LLMs, which tend to avoid directly assigning the 'Fatal accident' label due to their training to steer clear of discussing unpleasant or unsafe topics related to human death. With comparison to other settings, this could highlight whether prompt engineering enhances LLMs' performance in traffic safety analysis by addressing inherent biases and improving the model's ability to more reliably infer the fatal outcome of traffic incidents.

\begin{figure}[H]
    \centering
    \includegraphics[width=0.8\linewidth]{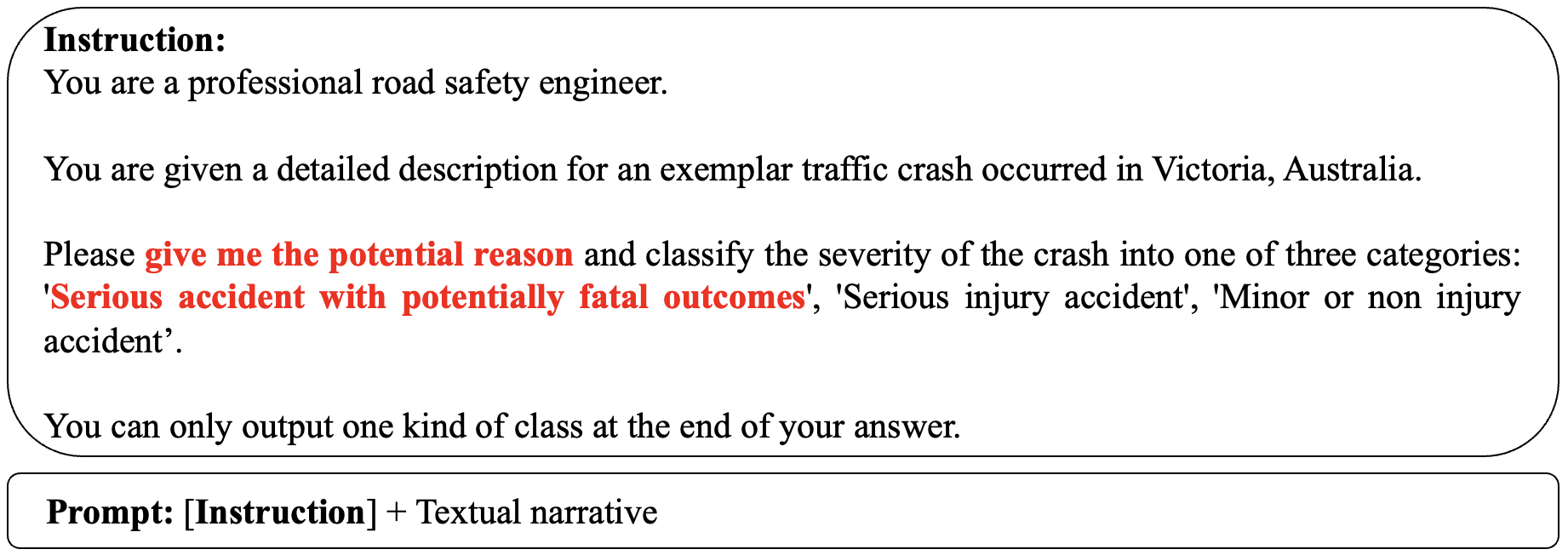}
    \caption{Zero-shot with prompt engineering \& CoT (ZS\_PE\_CoT)}
    \label{fig:ZSPECoT}
\end{figure}

The prompt designed for Zero-shot with PE \& CoT setting is shown in Figure \ref{fig:ZSPECoT}. In this prompt design, we not only included the CoT, by requiring a logical deduction from cause reasoning to severity outcome classification, but also changed the class label of 'Fatal accident' to the soft version of 'Serious accident with potentially fatal outcomes'. The differences between ZS\_PE\_CoT and plain ZS are highlighted in red in Figure \ref{fig:ZSPECoT}.

\subsubsection{Few-shot}

\begin{figure}[htbp]
    \centering
    \includegraphics[width=0.9\linewidth]{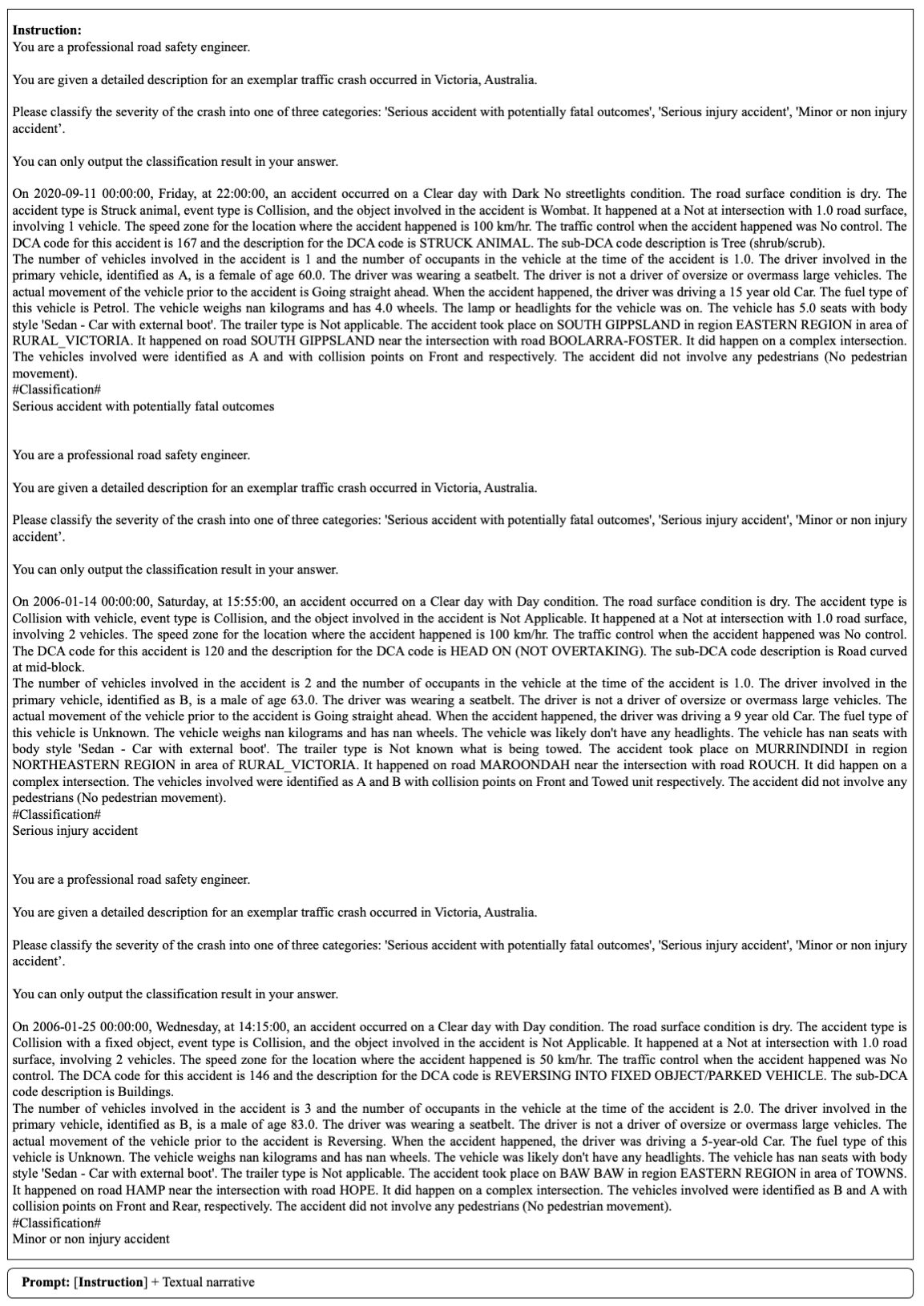}
    \caption{Few shot (FS)}
    \label{fig:Fewshot}
\end{figure}

The prompt designed for the plain few-shot setting is shown in Figure \ref{fig:Fewshot}. In this paper, the few-shot setting refers to a three-shot scenario, where three examples, one from each severity category, are provided for the LLMs to infer from. 

The only difference between the prompt for Few-shot with Prompt Engineering (FS\_PE) and that of the plain few-shot is that we substitute "Fatal accidents" with "Serous accidents with potentially fatal outcomes".

\subsection{Evaluation metrics}
Following standard practice in the context of multi-class classification, we adopt two commonly used classification metrics: Macro-Accuracy, and F1-score. Additionally, we include class-specific accuracies. These metrics are briefly discussed below.

(1) Accuracy

Accuracy measures the proportion of correctly classified instances in the test dataset. It is calculated as:

\begin{equation}
\text{Accuracy} = \frac{\text{Correct Predictions}}{\text{Total Predictions}}
\end{equation}

where:
\begin{itemize}
    \item $\text{Correct Predictions}$: The number of correctly classified instances in the test dataset.
    \item $\text{Total Predictions}$: The total number of instances in the test dataset.
\end{itemize}
It should be noted that we first calculate the accuracy for each class and then calculate the macro-accuracy as the average of these class accuracies. 

(2) F1-score

The F1-score is defined as the harmonic mean of precision and recall, computed as:

\begin{equation}
\text{F1-score} = \frac{2 \times \text{Precision} \times \text{Recall}}{\text{Precision} + \text{Recall}}
\end{equation}
The F1-score reported in the following section (Section 5) is at the macro level, which is an averaged F1 of all classes.

Precision quantifies the accuracy of positive predictions for a specific class, computed as:

\begin{equation}
\text{Precision} = \frac{\text{True Positives}}{\text{True Positives} + \text{False Positives}}
\end{equation}

where:
\begin{itemize}
    \item $\text{True Positives}$: The number of correctly predicted instances of the class.
    \item $\text{False Positives}$: The number of instances wrongly classified into the class.
\end{itemize}

Recall, also known as sensitivity or true positive rate, measures the ability of the model to correctly identify instances of a specific class. It is calculated as:

\begin{equation}
\text{Recall} = \frac{\text{True Positives}}{\text{True Positives} + \text{False Negatives}}
\end{equation}

where:
\begin{itemize}
    \item $\text{False Negatives}$: The number of instances of the class wrongly classified as something else.
\end{itemize}

\section{Findings}
\subsection{Exemplar responses of LLMs to crash severity inference queries}

\begin{figure}[htbp]
    \centering
    \includegraphics[width=1\linewidth]{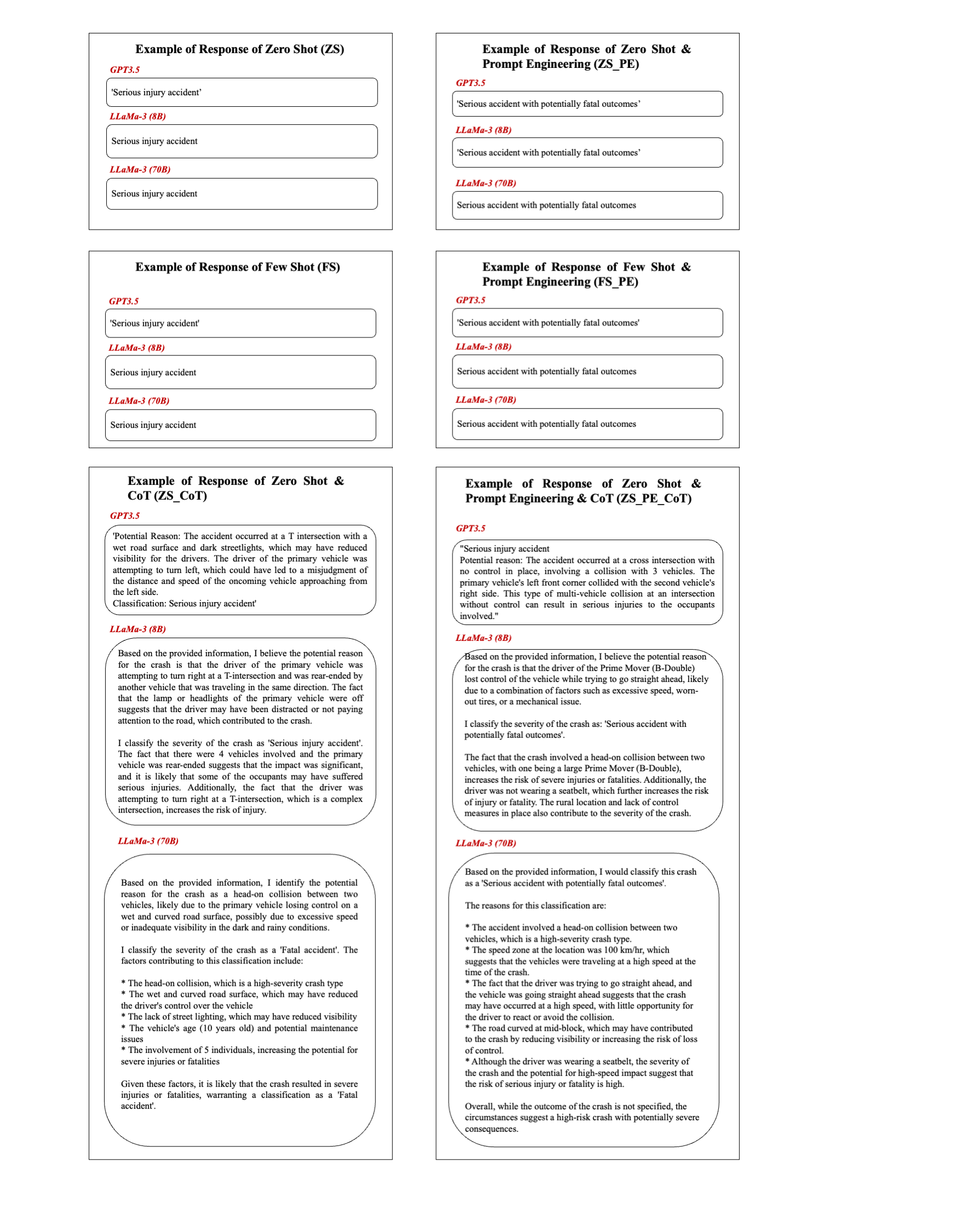}
    \caption{Exemplar responses of LLMs in different settings}
    \label{fig:responses}
\end{figure}

The exemplar responses of GPT-3.5, LLaMA3-8B, and LLaMA3-70B in each of the six settings, including ZS, FS, ZS\_CoT, FS\_PE, ZS\_PE, and ZS\_PE\_CoT (outlined in Table \ref{experiments}), are shown in Figure \ref{fig:responses}. It demonstrates that the LLMs can effectively respond to the severity inference task, delivering expected results. Note that the examples in Figure \ref{fig:responses} only showcase correct severity inferences.

Given the prompt for each setting (see Section 4.2), each model can directly answer or ultimately summarize its estimated severity for the given accident as one of the defined categories.

In the plain Zero-shot and Few-shot settings, the models respond directly with one of the three class labels, i.e., 'Minor or non-injury accident', 'Serious injury accident', or 'Fatal accident'. Similarly, in the ZS\_PE and FS\_PE settings, the models respond directly as 'Minor or non-injury accident', 'Serious injury accident', or 'Serious accident with potentially fatal outcomes'.

In contrast, in the CoT settings (ZS\_CoT and ZS\_PE\_CoT), the models return longer responses by reasoning first and then making inference of the severity outcome of the accident. Generally, the GPT-3.5 model's responses are more concise.

\subsection{Severity inference performance of the LLMs with different strategies}

The performance metrics of GPT-3.5, LLaMA3-8B, and LLaMA3-70B for the crash severity inference task under the six settings (see Table 2) are presented in Table \ref{tab:performance}. 

The results reveal varied performance across models and settings in inferring crash severity outcomes. LLaMA3-70B consistently exhibited superior performance, particularly with zero-shot prompt engineering (ZS\_PE), achieving the highest macro F1-score (0.4755) and macro-accuracy (49.33\%). Furthermore, LLaMA3-70B attained the second-best performance in macro F1-score (0.4747) and macro-accuracy (47.33\%) under the zero-shot with Chain-of-Thought (ZS\_CoT) setting. These findings suggest that both prompt engineering and Chain-of-Thought methodologies contribute positively to model performance. Nevertheless, no single technique demonstrated consistent superiority across all severity categories. For fatal accidents, GPT-3.5 with zero-shot prompt engineering and Chain-of-Thought (ZS\_PE\_CoT) exhibited the highest accuracy (68\%). In contrast, for serious injury accidents, GPT-3.5 and LLaMA3-8B in the plain zero-shot setting (ZS), as well as GPT-3.5 in the zero-shot with Chain-of-Thought scenario (ZS\_CoT), achieved 100\% accuracy. However, it is crucial to note that in these settings, these models performed poorly for fatal and 'minor or non-injury' accidents, indicating an inherent bias toward the intermediate severity category of 'serious injury'.

Interestingly, LLaMA3-70B with the basic zero-shot approach demonstrated the best inference performance for 'minor or non-injury' accidents (58\% accuracy) while maintaining a relatively balanced performance across fatal and serious injury accidents. This suggests a robust generalization capability of LLaMA3-70B across crash severity categories.

The implementation of prompt engineering, particularly in zero-shot settings, generally enhanced performance across models. This improvement was especially pronounced for fatal accident classification, where rephrasing the "Fatal accidents" label to the soft version of "Serious accident with potentially fatal outcomes" facilitated maintenance of classification accuracy while adhering to the LLM's aligned behaviors.

These results underscore the complexity inherent in crash severity inference task, as no single approach consistently outperformed others across all metrics and severity categories. These findings highlight the need for careful selection of models and methodologies based on specific task requirements and the relative importance of different severity categories in the application context.

\begin{table}[htbp]
\centering
\caption{Performance of Models on Crash Severity Inference Task}
\label{tab:performance}
\begin{adjustbox}{max width=\textwidth}
\begin{tabular}{llcccccc}
\toprule
 & \textbf{Model} & \textbf{Macro F1-score} & \textbf{Macro-accuracy} & \textbf{Fatal accident} & \textbf{Serious injury accident} & \textbf{Minor or non-injury accident} \\
\midrule
\multicolumn{2}{l}{\textbf{\textit{ZS}}} \\
 & GPT-3.5 & 0.1812 & 0.3400 & 0.00 & \textbf{1.00} & 0.02 \\
 & LLaMA3-8B & 0.1818 & 0.3400 & 0.00 & \textbf{1.00} & 0.02 \\
 & LLaMA3-70B & 0.4541 & 0.4533 & 0.44 & 0.34 & \textbf{0.58} \\
\multicolumn{2}{l}{\textbf{\textit{ZS\_CoT}}} \\
 & GPT-3.5 & 0.2073 & 0.3533 & 0.00 & \textbf{1.00} & 0.06 \\
 & LLaMA3-8B & 0.2496 & 0.3533 & 0.00 & 0.88 & 0.18 \\
 & LLaMA3-70B & \underline{0.4747} & \underline{0.4733} & 0.40 & 0.64 & \underline{0.38} \\
\multicolumn{2}{l}{\textbf{\textit{ZS\_PE}}} \\
 & GPT-3.5 & 0.3798 & 0.4533 & \underline{0.62} & 0.72 & 0.02 \\
 & LLaMA3-8B & 0.3120 & 0.4000 & 0.34 & 0.86 & 0.00 \\
 & LLaMA3-70B &\textbf{ 0.4755} & \textbf{0.4933} & 0.60 & 0.66 & 0.22 \\
\multicolumn{2}{l}{\textbf{\textit{ZS\_PE\_CoT}}} \\
 & GPT-3.5 & 0.3509 & 0.4200 & \textbf{0.68} & 0.56 & 0.02 \\
 & LLaMA3-8B & 0.4033 & 0.4533 & 0.60 & 0.68 & 0.08 \\
 & LLaMA3-70B & 0.3581 & 0.4267 & \underline{ 0.62} & 0.64 & 0.02 \\
\multicolumn{2}{l}{\textbf{\textit{FS}}} \\
 & GPT-3.5 & 0.2514 & 0.3667 & 0.04 & 0.96 & 0.10 \\
 & LLaMA3-8B & 0.4068 & 0.4267 & 0.22 & 0.72 & 0.34 \\
 & LLaMA3-70B & 0.4131 & 0.4200 & 0.26 & 0.64 & 0.36 \\
\multicolumn{2}{l}{\textbf{\textit{FS\_PE}}} \\
 & GPT-3.5 & 0.2576 & 0.3667 & 0.18 & 0.92 & 0.00 \\
 & LLaMA3-8B & 0.2928 & 0.3933 & 0.08 & \underline{0.98} & 0.12 \\
 & LLaMA3-70B & 0.3856 & 0.4600 & 0.56 & 0.80 & 0.02 \\
\bottomrule

\multicolumn{6}{l}{{\itshape --\textbf{Bold} values are the best per column.}}\\
\multicolumn{6}{l}{{\itshape --\underline{underlined} values are the second best per column.}}
\end{tabular}
\end{adjustbox}
\end{table}

\subsection{Effectiveness of prompt engineering (PE) and Chain-of-Thought (CoT)}

\begin{figure}[htbp]
    \centering
    \includegraphics[width=0.9\linewidth]{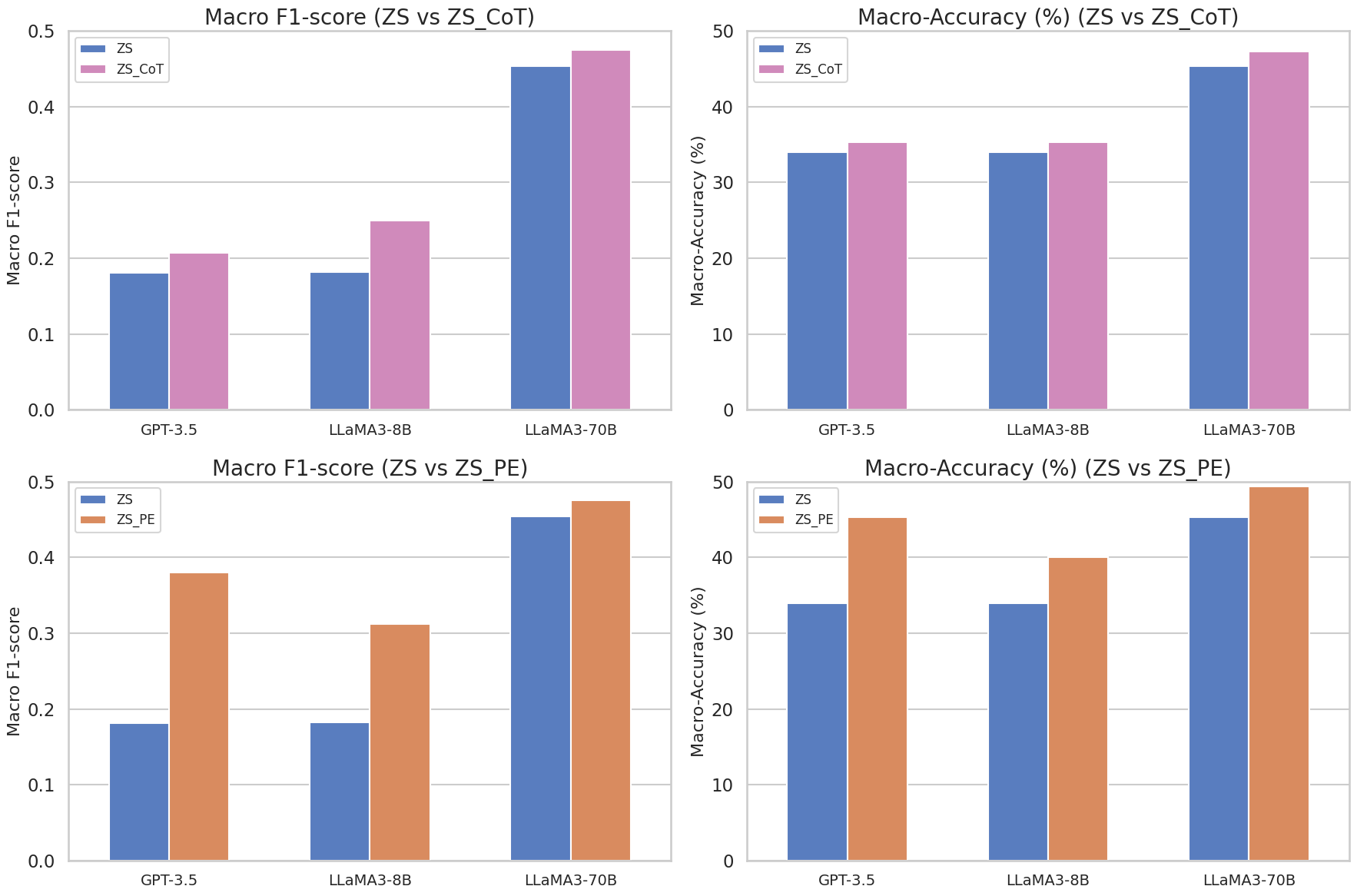}
    \caption{Effect of PE or CoT separately }
    \label{fig:zscotpe}
\end{figure}

Figure \ref{fig:zscotpe} demonstrates the performance gains by CoT and PE as compared to the plain zero-shot setting. Both ZS\_CoT and ZS\_PE consistently demonstrate enhanced performance in terms of Macro F1-score and Macro-accuracy across all three models evaluated. This improvement underscores the efficacy of CoT and PE in boosting model performance in zero-shot scenarios.

Notably, the implementation of PE (ZS\_PE) yields more substantial improvements relative to CoT (ZS\_CoT). This differential in enhancement suggests that, within the context of this specific task, the reformulation of prompts may be particularly effective in guiding model outputs as compared to the structured reasoning approach with CoT. The consistent pattern of improvement across different model architectures and sizes indicates the broad applicability of these techniques in zero-shot learning paradigms. 


As illustrated in Figure \ref{fig:zscotpe}, CoT improves both Macro F1-score and Macro-Accuracy across all three models in the plain zero-shot (ZS) setting. Based on the results summarized in Table \ref{tab:performance}, GPT-3.5 and LLaMA3-8B show improved recognition of "Minor or non-injury" accidents. LLaMA3-70B demonstrates substantial gain in identifying "Serious injury" accidents and "Minor or non-injury" accidents, with only a slight reduction in performance for "Fatal" accidents. The use of CoT enables LLMs to better understand and reason through questions, leading to more reliable and explainable inferences. 

The PE technique also leads to increased Macro F1-score and Macro-Accuracy across all three models compared to the plain zero-shot (ZS) setting, as depicted in Figure \ref{fig:zscotpe}. Notably, it greatly enhances the models' ability to detect Fatal accidents by simply softening the label description from "Fatal accident" to "Serious accident with potentially fatal outcomes", resulting in more balanced performance across severity categories. Compared to the zero-shot baseline, GPT-3.5 with PE attains a remarkable increase in Fatal accident detection, with accuracy rising  from 0\% to 62\%. Similarly, LLaMA3-8B and LLaMA3-70B show increases in fatal accident accuracy from 0\% to 34\%, 44\% to 60\%, respectively. 

These improvements may stem from the fact that PE directs LLMs to concentrate more specifically on accident severity classification, potentially addressing any initial tendency to be overly cautious or generalized in their responses. This targeted guidance enables the models to make more precise distinctions among accident severity categories.

\begin{figure}[H]
    \centering
    \includegraphics[width=1\linewidth]{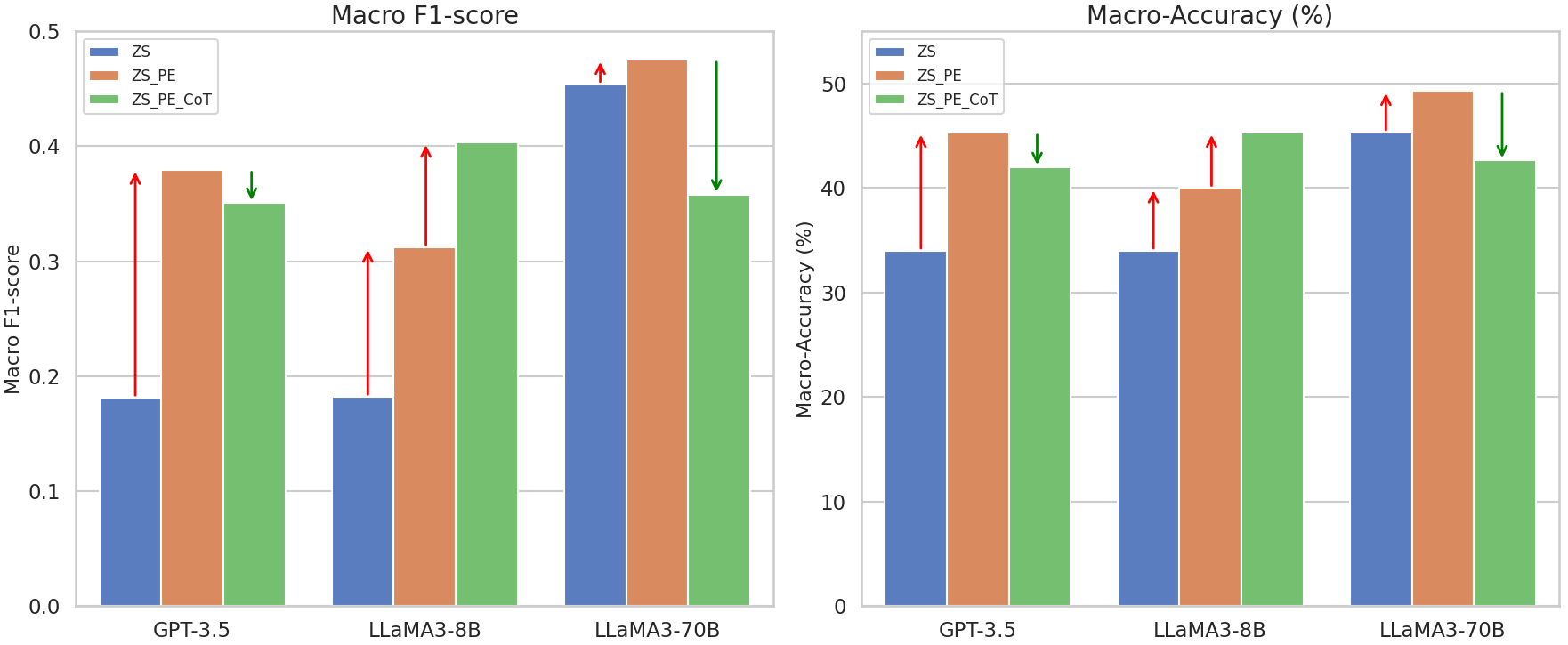}
    \caption{Performance comparison of models in ZS, ZS\_PE, and ZS\_PE\_CoT}
    \label{fig:pecot}
\end{figure}

Figure \ref{fig:pecot} shows the comparative performance of the three models across incremental settings from ZS to ZS\_PE and ZS\_PE\_CoT. 

In both the zero-shot (ZS) and zero-shot with PE (ZS\_PE) settings, LLaMA3-70B consistently outperforms the other two models. In the ZS setting, LLaMA3-70B achieves a macro F1-score of 0.4541 and a macro-accuracy of 45.33\%, significantly higher than both GPT-3.5 (0.1812 and 34.00\%) and LLaMA3-8B (0.1818 and 34.00\%). This performance advantage is maintained in the ZS\_PE setting, where LLaMA3-70B shows further improvement with a macro F1-score of 0.4755 and a macro-accuracy of 49.33\%, compared to GPT-3.5 (0.3798 and 45.33\%) and LLaMA3-8B (0.3120 and 40.00\%).

However, the performance dynamics shift in the setting of zero-shot with PE and CoT (ZS\_PE\_CoT). In this setting, LLaMA3-8B leads the performance with a macro F1-score of 0.4033 and a macro-accuracy of 45.33\%, surpassing both GPT-3.5 (0.3509 and 42.00\%) and LLaMA3-70B (0.3581 and 42.67\%). GPT-3.5 and LLaMA3-70B experience slightly decreased performance under the ZS\_PE\_CoT setting as compared to the ZS\_PE setting. In contrast, LLaMA3-8B show improved macro-F1-score from 0.3120 to 0.4033 as well as increased macro-accuracy from 0.4000 to 0.4533. This shift in performance suggests that the combination of PE and CoT reasoning are particularly more beneficial to small models, such as LLaMA3-8B, than large models like GPT-3.5 and LLaMA3-70B for the crash severity inference task. 

In the ZS\_PE\_CoT setting, all three models, GPT-3.5, LLaMA3-8B, and LLaMA3-70B, demonstrate improved recognition of fatal accidents, evidenced in Table \ref{tab:performance}. This enhancement indicates that the combination of CoT and PE is particularly beneficial for identifying more severe crashes than less severe ones.

\subsection{Zero-shot vs. Few-shot learning}

In the FS setting, the inclusion of three examples improves both the macro F1-score and macro-accuracy compared to the ZS setting, boosting classification accuracy of  GPT-3.5 and LLaMA3-8B for “Fatal accident” and “Minor or non-injury accident”. However, this comes at the expense of accuracy for “Serious injury accident”. This indicates a potential trade-off in classification performance across severity categories or a decrease of the model bias in the ZS setting. 

Moreover, smaller models like LLaMA3-8B generally benefit more from few-shot learning than larger models, such as GPT-3.5 and LLaMA3-70B, as evidenced by a notable increase in Macro F1 score from 0.1818 to 0.4068. Nevertheless, LLaMA3-70B, being a larger model, performs slightly better in the zero-shot settings, suggesting it may have gained some general knowledge in traffic safety domain during the pre-training stage, where the zero-shot prompting can draw upon such knowledge.


In contrast, the effects of PE in the FS setting exhibit more variations across models. GPT-3.5 demonstrates improvements in Macro F1-score and Fatal accident accuracy, and LLaMA3-70B shows remarkably improved inference accuracy for "fatal accident" and "Serious injury accident". Conversely, LLaMA3-8B shows decreased macro F1-score and macro-accuracy, indicating that PEin the few-shot setting may not be equally beneficial for models of different sizes. 

It important to note that we did not explore the aspect regarding the choice of the examples in the FS setting, which might have a varying effect on different models.

\section{Discussions}

\subsection{Can LLMs with CoT yield logical reasoning for their inference outcomes?}

CoT is a unique technique to augment LLM's reasoning capability. But how reasonable is the reasoning by LLMs with CoT? We aim to examine the responses of LLMs with CoT from the view of a traffic safety engineer. Specifically, the responses of LLaMA3-70B in the ZS\_CoT setting are evaluated due to its best performance in this setting (see Table \ref{tab:performance}).

\begin{figure}[ht]
    \centering
    \includegraphics[width=0.85\linewidth]{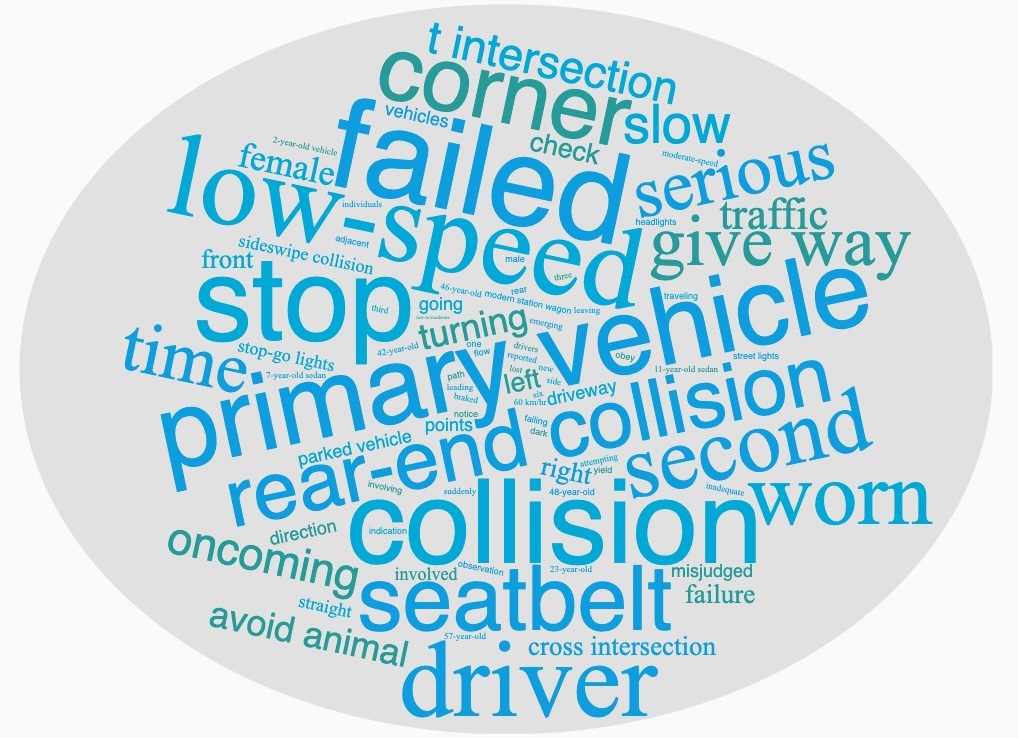}
    \caption{Word cloud for correctly inferred "Minor or non-injury accident" in the ZS\_CoT setting}
    \label{fig:minor_wordcloud}
\end{figure}

\begin{figure}[ht]
    \centering
    \includegraphics[width=0.85\linewidth]{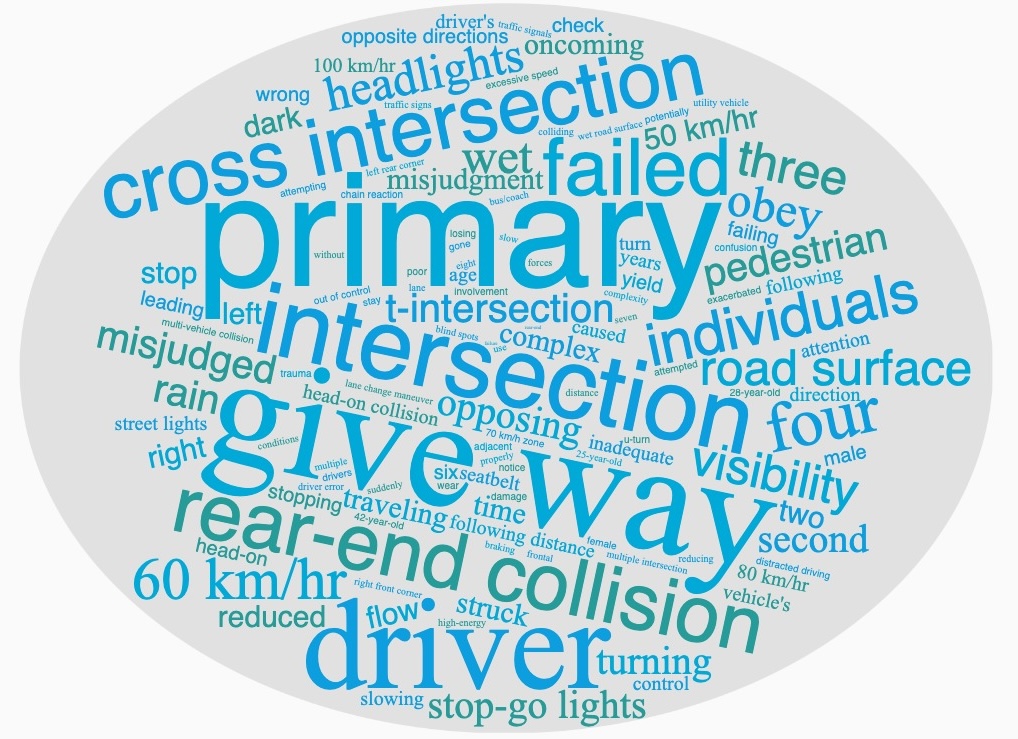}
    \caption{Word cloud for correctly inferred "Serious injury accident" in the ZS\_CoT setting}
    \label{fig:serious_wordcloud}
\end{figure}

\begin{figure}[ht]
    \centering
    \includegraphics[width=0.85\linewidth]{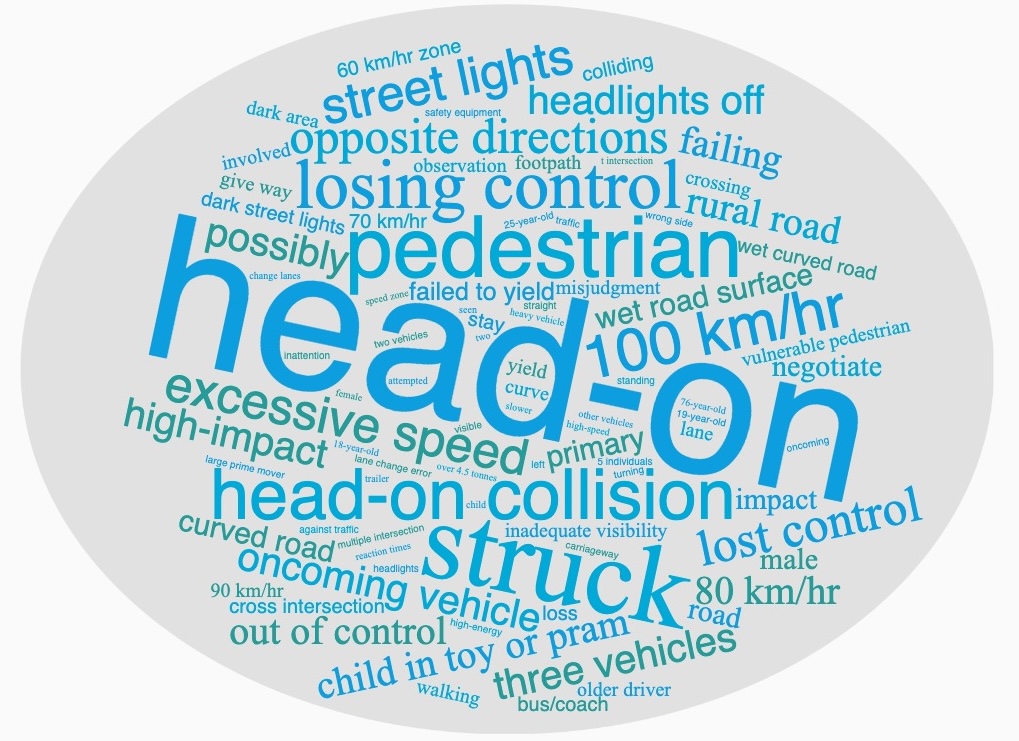}
    \caption{Word cloud for correctly inferred "Fatal accident" in the ZS\_CoT setting}
    \label{fig:fatal_wordcloud}
\end{figure}

As a qualitative assessment, three word clouds are drawn separately with respect to the three severity categories, i.e., "Minor/non-injury", "Serious injury", and "Fatal" accidents (see Figures \ref{fig:minor_wordcloud}, \ref{fig:serious_wordcloud},and \ref{fig:fatal_wordcloud}. Note that only the correct inferred responses are used in creating these word clouds, where the bigger sizes of words indicate their higher frequencies in the LLMs' responses. These visualizations offer insights into the LLM's conceptualization and reasoning processes regarding accident causation and factors considered during the severity inference.

In the three word clouds (Figure \ref{fig:minor_wordcloud}, \ref{fig:serious_wordcloud},and \ref{fig:fatal_wordcloud}), some words consistently appear regardless of severity outcomes, including "collision," "intersection," "vehicle," and "driver." This suggests a core set of concepts that the LLM associates with traffic accidents. Additionally, LLaMA3-70B demonstrates consideration of diverse factors in its accident analysis, including:

$\bullet$ Crash-related factors (e.g., "rear-end collision", "pedestrian", "opposite directions", "corner")

$\bullet$ Environmental conditions (e.g., "wet road surface", "rain", "dark", "stop-go")

$\bullet$ Driver behavior (e.g., "failing to yield," "misjudgment", "turning", "give way", "excessive speed")

$\bullet$ Driver characteristics (e.g., "male" "older driver", "age")

$\bullet$ Vehicle factors (e.g., "bus", "headlights," "seatbelt")

$\bullet$ Road design elements (e.g., "traffic lights","intersection," "curved road", "t intersection")

The word cloud for "Minor or non-injury" accidents (Figure \ref{fig:minor_wordcloud}), is characterized by distinct terms suggesting lower impact and preventative measures, such as "Low speed" and "slow", indicating the LLM associates lower velocities with less severe outcomes.
"Seatbelt" and "failed" suggest a focus on minor infractions and safety equipment.
"Stop" and "time" may relate to issues at intersections or reaction times.
"Rear-end collision", and "corner" also are prominent, indicating that LLM associates rear-end collision or accident impact on corners of vehicles with less severe outcomes. 

The word cloud for "Serious injury accident", shown in Figure \ref{fig:serious_wordcloud}, is characterized by terms suggesting moderate impact, such as those relating to drivers (e.g., "give way", "misjudged") and locations (e.g., "intersection", "cross intersection"). "Wet," "rain," and "visibility" emerge, highlighting adverse weather or environmental conditions. "60 km/hr" and "50 km/hr" reveals relatively high speeds are associated with more severe outcomes. "Four" and "three" appear, indicating potentially more vehicles or individuals involved in an accident with severe outcomes. "Rear-end collision" appears again, which combines with "intersections" and the higher speeds (e.g., "60 km/hr"), could lead to "Serious injury" accidents.

\begin{figure}
    \centering
    \includegraphics[width=1\linewidth]{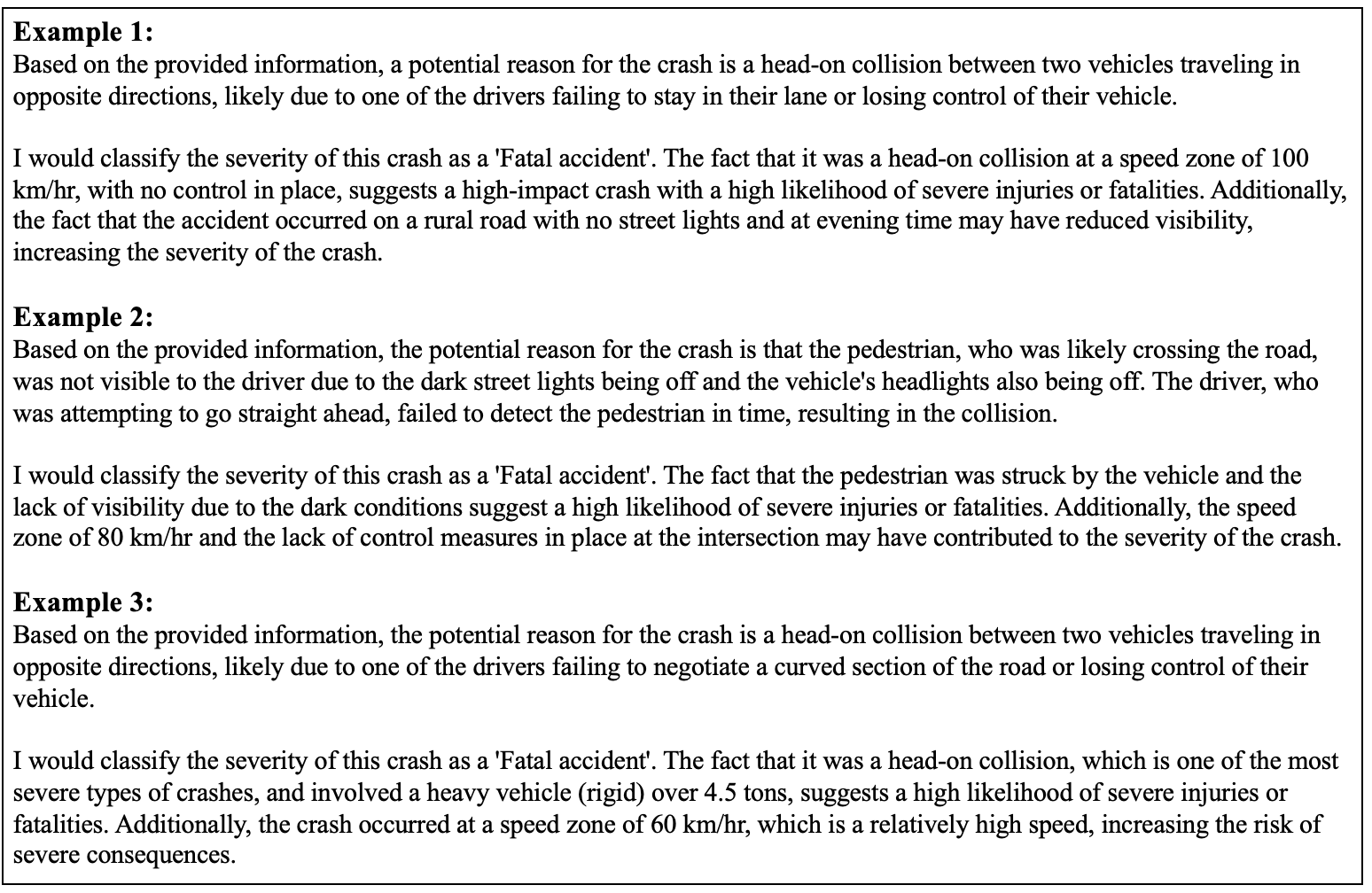}
    \caption{Output examples for fatal accidents from LLaMA3-70B in ZS\_CoT setting}
    \label{fig:fatal_output_cot}
\end{figure}

The word cloud for "Fatal accident" reveals a marked shift towards high-energy impact scenarios (see Figure \ref{fig:fatal_wordcloud}). "Head-on" and "high-impact" are among the most prominent terms. Specifically, higher speeds and speeding (e.g., "100 km/hr", "excessive speed") are associated with fatal accidents. 
"Pedestrian" gains significance, reflecting heightened vulnerability. "Losing control" and "out of control" suggest more catastrophic situations or driver errors. "Rural road", "curved road" are also important terms identified in the word cloud. Furthermore, four distinct examples are provided in Figure \ref{fig:fatal_output_cot} to illustrate the CoT's reasoning process.

\subsection{Limitations and future works}
One of the primary limitations of this study is the relatively small sample used, including only 150 instances (50 for each severity category). This limited dataset may not fully capture the variability and complexity of diverse real-world crash scenarios, potentially affecting the generalizability of the findings.

For future research, several directions can be explored to further enhance the performance and applicability of LLMs in crash analysis and modeling:

1) Expanding the dataset to include a larger and more diverse set of samples will allow for a more comprehensive evaluation of the models' capabilities and improve the robustness of the results.

2) Fine-tuning LLMs with more extensive and domain-specific data (e.g., crash reports and databases) can significantly enhance their domain knowledge to better understand the nuances and specificities of traffic accidents, leading to more accurate and reliable inference.

3) Investigating explanation methods in conjunction with LLMs can yield more interpretable and trustworthy results.

\section{Conclusions}

In conclusion, this study demonstrates the efficacy of LLMs in crash severity inference using textual narratives of crash events constructed from structured tabular data. Our comprehensive evaluation of modern LLMs (GPT-3.5-turbo, LLaMA3-8B, and LLaMA3-70B) across different settings (zero-shot, few-shot, CoT, and PE) yields insightful findings. LLaMA3-70B consistently outperformed other models, especially in zero-shot settings. CoT and PE techniques lead to enhanced performance, improving logical reasoning and addressing alignment issues.

Notably, the use of CoT provided valuable insights into LLM reasoning processes, revealing their capacity to consider multiple factors such as environmental conditions, driver behavior, and vehicle characteristics in the crash severity inference task. These findings collectively suggest that LLMs hold considerable promise for crash analysis and modeling. Future research may explore other safety applications beyond the severity inference.

\section{Author contributions}
The authors confirm contribution to the paper as follows: study conception and design: J. Yang, H. Zhen, N, Liu; data processing and cleaning: H. Zhen, Y. Shi; experiments design, analysis and interpretation of results: H. Zhen, Y. Shi, J. Yang; draft manuscript preparation: H. Zhen, Y. Shi, Y. Huang; review and editing: J. Yang, N. Liu. All authors reviewed the results and approved the final version of the manuscript.

\bibliographystyle{plainnat}  
\bibliography{references}

\end{document}